\documentclass[10pt,twocolumn,letterpaper]{article}

\usepackage{iccv}
\usepackage{times}
\usepackage{epsfig}
\usepackage{graphicx}
\usepackage{amsmath}
\usepackage{amssymb}

\usepackage{url}            % simple URL typesetting
\usepackage{amsfonts}       % blackboard math symbols
\usepackage{nicefrac}       % compact symbols for 1/2, etc.
\usepackage{color}

\usepackage{bm}
\usepackage{multirow}
\usepackage{tabularx}
\usepackage{booktabs}
\usepackage[ruled,vlined]{algorithm2e}
\usepackage{algorithmic}
\usepackage{float}
\usepackage{wrapfig}
\usepackage{textcomp}

\newcommand{\hide}[1]{}

\newcommand{\citep}{\cite}
\newcommand{\citet}{\cite}

\usepackage[breaklinks=true,bookmarks=false]{hyperref}

\iccvfinalcopy % *** Uncomment this line for the final submission

\def\iccvPaperID{4634} % *** Enter the ICCV Paper ID here

\ificcvfinal\fi

\begin{document}

%%%%%%%%% TITLE
\title{Bayesian Relational Memory for Semantic Visual Navigation}

\author{
\begin{tabular}{cccccc}
    Yi Wu\footnotemark[2]~\;\footnotemark[5]\;\;\,&Yuxin Wu\footnotemark[3]\;\;\,& Aviv Tamar\footnotemark[4] \;\;\,& Stuart Russell\footnotemark[2] \;\;\,& Georgia Gkioxari\footnotemark[3]\;\; \,& Yuandong Tian\footnotemark[3]
\end{tabular}\\
\begin{tabular}{cccc}
   \footnotemark[2]\,\;UC Berkeley\;\;&\footnotemark[3]\,\;Facebook AI Research\;\;&\footnotemark[4]\,\;Technion\;\;&\footnotemark[5]\,\;OpenAI
\end{tabular}\\
\begin{tabular}{ccc}
    \small{\footnotemark[2]}\,\;\footnotesize{\texttt{\{jxwuyi,russell\}@cs.berkeley.edu}} \;\;& \small{\footnotemark[3]}\,\;\footnotesize{\texttt{\{yuxinwu,gkioxari,yuandong\}@fb.com}}
    &\small{\footnotemark[4]}\,\;\footnotesize{\texttt{avivt@technion.ac.il}}
\end{tabular}
}

\maketitle
\ificcvfinal\thispagestyle{empty}\fi

%%%%%%%%% ABSTRACT
\begin{abstract}
We introduce a new memory architecture, \emph{Bayesian Relational Memory} (BRM), to improve the generalization ability for semantic visual navigation agents in unseen environments, where an agent is given a semantic target to navigate towards. BRM takes the form of a probabilistic relation graph  over semantic entities (e.g., room types), which allows (1) capturing the layout prior from training environments, i.e., \emph{prior knowledge}, (2) estimating posterior layout  at test time, i.e., \emph{memory update}, and (3) efficient \emph{planning} for navigation, altogether. We develop a BRM agent consisting of a BRM module for producing sub-goals and a goal-conditioned locomotion module for control. When testing in unseen environments, the BRM agent outperforms baselines that do not explicitly utilize the probabilistic relational memory structure.
\end{abstract}
% When testing in unseen environments in House3D\yuandong{reference}

%%%%%%%%% BODY TEXT

\begin{figure*}[bt!]
    \centering
    \includegraphics[width=0.9\textwidth]{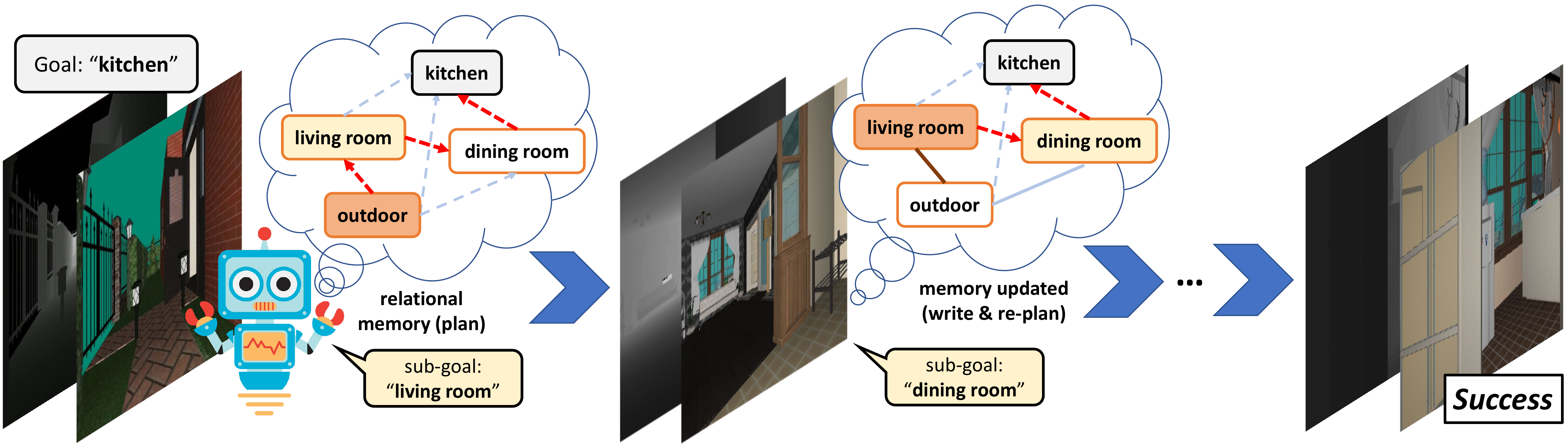}
    
    \caption{A demonstration of our task and method. The agent perceives visual signals and needs to find the kitchen, which cannot be seen from outside (leftmost). So the agent plans in its memory and conclude that it may first find a likely intermediate waypoint, i.e., living room. Then the agent repeatedly updates its belief of the environment layout, re-plans accordingly and reaches the kitchen in the end.}
    \label{fig:task}
    \vspace{-2mm}
\end{figure*}

\section{Introduction}

Memory is a crucial component for  intelligent agents to gain extensive reasoning abilities over a long horizon. One such challenge is visual navigation, where an agent is placed to an environment with unknown layouts and room connectivity, acts on visual signal perceived from its surrounding, explores and reaches a goal position efficiently. 

Due to partial observability, the agent must memorize its past experiences and react accordingly. Hence, deep learning (DL) models for visual navigation often encodes memory structures in its design. LSTM is initially used to as general-purpose implicit memory~\cite{mirowski2016learning,mirowski2018learning,das2017embodied}. Recently, to improve the performance, explicit and navigation-specific structural memory are used~\cite{parisotto2017neural,gupta2017cognitive,savinov2018graph,gupta2017unifying}. 

Two categories exist for navigation-specific memory: the spatial memory~\cite{parisotto2017neural,gupta2017cognitive} and topological memory~\cite{gupta2017unifying,savinov2018graph}. The core idea of spatial memory is to extend the 1-dimensional memory in LSTM to a 2-dimensional matrix that represents the spatial structure of the environment, where a particular entry in the matrix corresponds to a 2D location/region in the environment. Due to its regular structure, value iteration~\cite{tamar2016value} can be applied directly for effective planning over the memory matrix. 

However, planning on such spatial memory can be computationally expensive for environments with large rooms. To navigate, precise localization of an agent is often not unnecessary. Extensive psychological evidences~\cite{savinov2018graph} also show that animals do not rely strongly on metric representations~\cite{wang2002human,foo2005humans}. Instead, humans primarily depend on a landmark-based navigation strategy, which can be supported by qualitative topological knowledge of the environment~\cite{foo2005humans}. Therefore, it is reasonable to represent the memory as a topological graph where the vertices are landmarks in the environment and edges denote short-term reachability between landmarks. During navigation, a localization network is trained to identify the position of the agent and the goal w.r.t. the landmarks in the memory and an efficient graph search can be used for long-term planning. 

However, still human navigation shows superior generalization performance which cannot be explained by either spatial or topological memory. For example, first-time home visitors naturally move towards the kitchen (rather than outdoor or toilet) to get a plate; from kitchen to bedroom, they know that living room may be an intermediate waypoint. Although visually different, such \emph{semantic knowledge}, i.e., the ``close-by'' relations over semantic entities, are naturally shared across environments and can be learned from previous experience to guide future navigation. In comparison, existing approaches of topological memory assume pre-exploration experiences of the environment before navigation starts~\cite{gupta2017unifying,savinov2018graph}, provide no memory updating operations, and cannot incorporate the prior knowledge of scene layouts and configurations from previously seen environments.

\iffalse 
which humans always utilize in the real world to navigate in novel scenarios. However, such relational structure of the prior knowledge does not fit into the paradigm of spatial memory while the topological memory paradigm is specialized to a single test environment.
\fi

In this work, we propose a new memory design for visual navigation, Bayesian Relational Memory (BRM), which (1) captures the prior knowledge of scene layouts from training environments and (2) allows both efficient planning and updating during test-time exploration. BRM can be viewed as a probabilistic version of a topological memory with semantic abstractions: each node in BRM denotes a semantic concept (e.g., object category, room type, etc), which can be detected via a neural detector, and each edge denotes the relation between two concepts. In each environment, a single relation may be present or not. For each relation (edge), we can average its existences over all training environments and learn an existence probability to denote the \emph{prior knowledge} between two semantic concepts. During exploration at test time, we can incrementally observe the existences of relations within that particular test environment. Therefore, we can use these environment specific observations to update the probability of those relations in the memory via the Bayes rule to derive the \emph{posterior knowledge}. Additionally, we train a semantic-goal-conditioned LSTM locomotion policy for control via deep reinforcement learning (DRL), and by planning on the relation graph with posterior probabilities, the agent picks the next semantic sub-goal to navigate towards. 

We evaluate our BRM method in a semantic visual navigation task on top of the House3D environment~\cite{wu2018building,das2017embodied}, which provides a diverse set of objects, textures and human-designed indoor scenes. The semantic scenes and entities in House3D are fully labeled (with noise), which are natural for our BRM model. In the navigation task, the agent observes first-person visual signals and needs to navigate towards a particular room type. We utilize the room types as the semantic concepts (nodes) in BRM and the "close-by" relations as the edges. We evaluate on unseen environments with random initial locations and compare our learned model against other DRL-based approaches without the BRM representation. Experimental results show that the agent equipped with BRM can achieve the semantic goal with higher success rates and fewer navigation steps. %\yuandong{Compare to what?}

Our contributions are as follows: (1) we proposed a new memory representation, Bayesian Relational Memory (BRM), in the form of probabilistic relation graphs over semantic concepts; (2) BRM is capable of encoding prior knowledge over training environments as well as efficient planning and updating at test time; (3) by integrating BRM into a DRL locomotion policy, we show that the generalization performances can be significantly improved.

\begin{figure*}[bt!]
    \centering
    \includegraphics[width=0.6\textwidth]{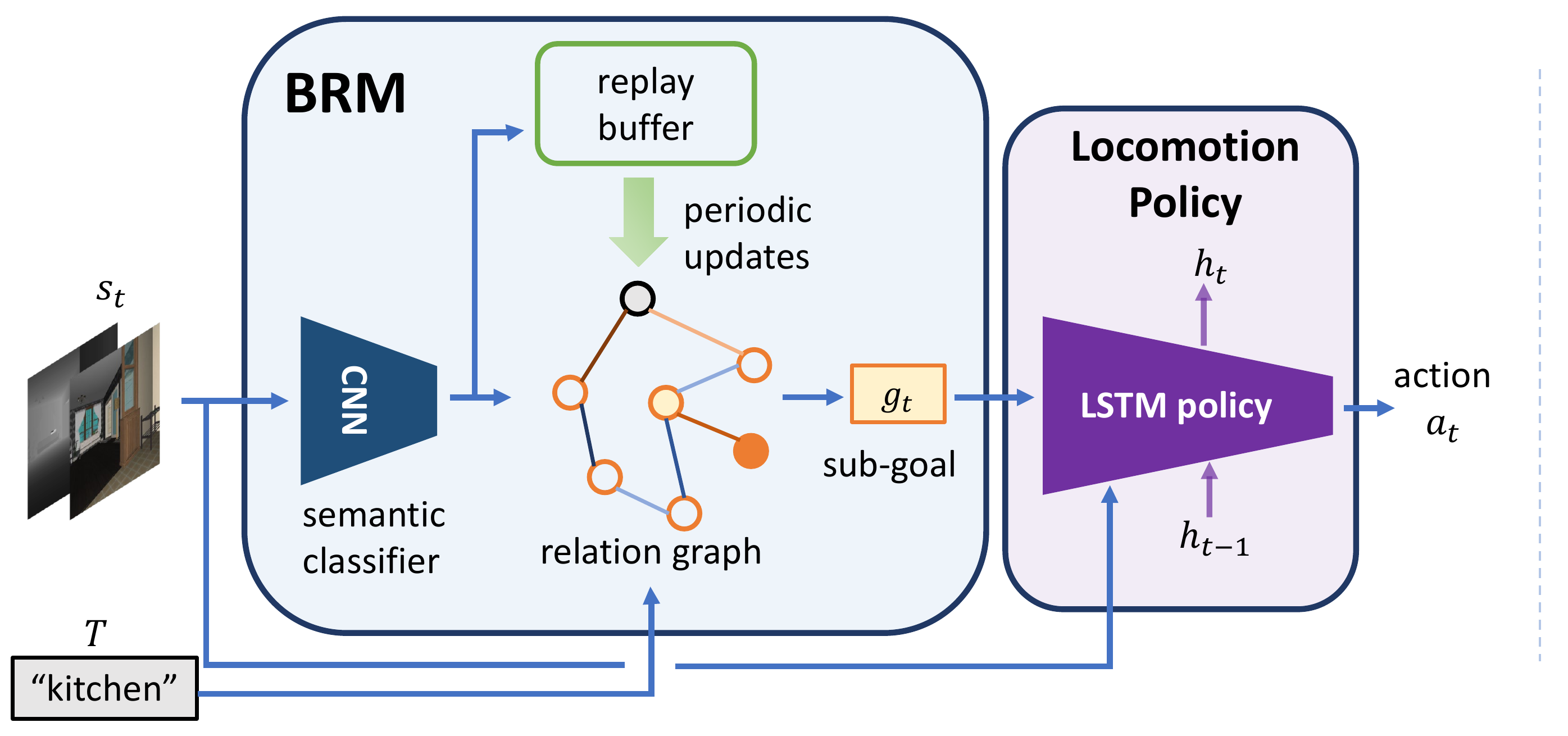}
    \includegraphics[width=0.2\textwidth]{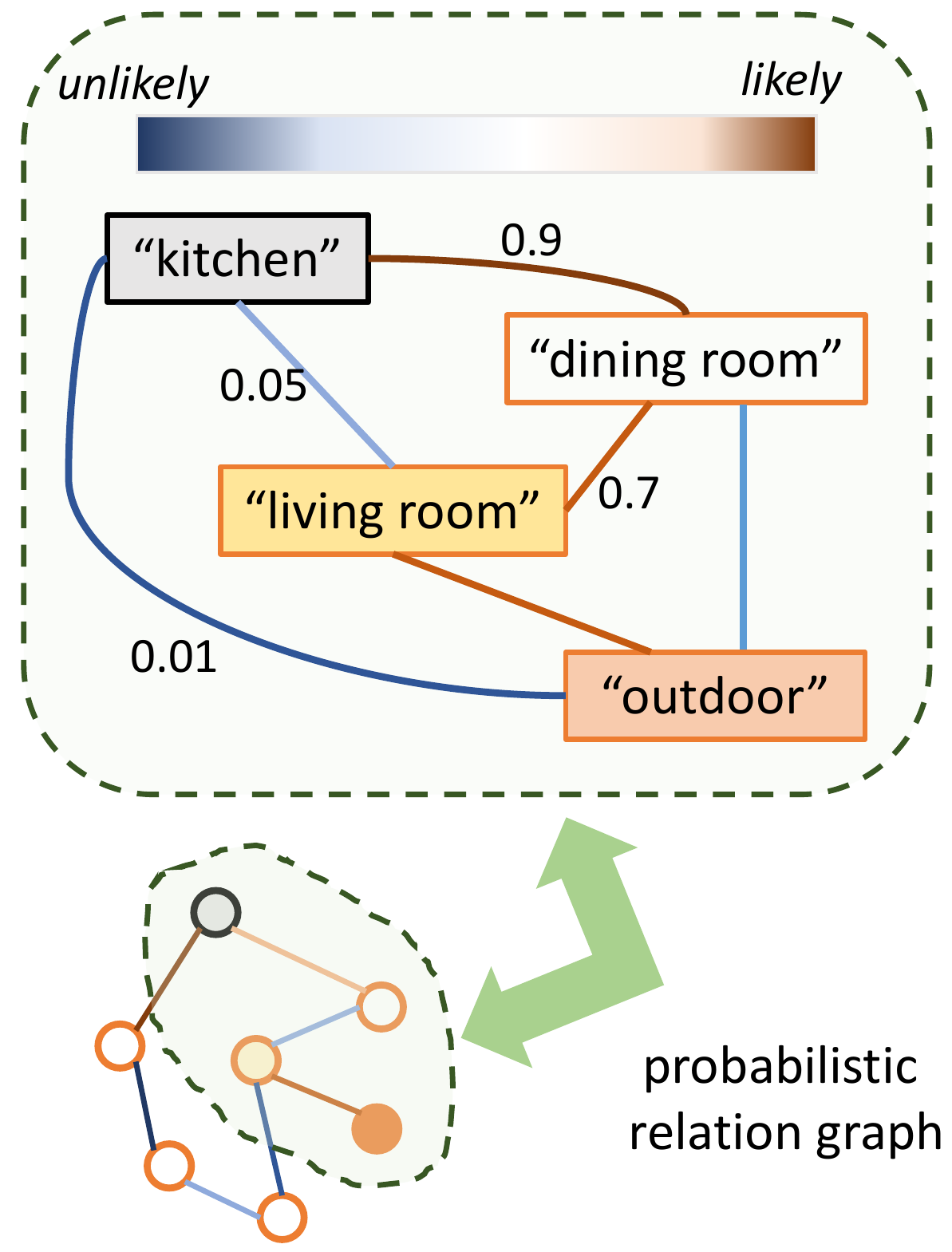}
    
    \caption{The architecture overview of our proposed navigation agent equipped with the Bayesian Relational Memory (BRM).}
    \label{fig:model}
    \vspace{-2mm}
\end{figure*}

\section{Related Work}

Navigation is one of the most fundamental problems in
mobile robotics. Traditional approaches (like SLAM) build metric maps via sensory signals, which is subsequently used for planning~\cite{elfes1987sonar,bonin2008visual,thrun2005probabilistic,durrant2006simultaneous}. More recently, thanks to the advances of deep learning, end-to-end approaches have been applied to tackle navigation in various domains, such  as mazes~\cite{mirowski2016learning,jaderberg2016reinforcement}, indoor scenes~\cite{zhu2017target,chang2017matterport3d,savva2017minos,mishkin2019benchmarking,kolve2017ai2}, autonomous driving~\cite{Chen_2015_ICCV,xu2017end,dosovitskiy2017carla}, and Google street view~\cite{mirowski2018learning}. There are also nice summaries of recent progresses~\cite{anderson2018evaluation,mishkin2019benchmarking}. We focus on indoor navigation scenario with House3D environment~\cite{wu2018building} which contains real-world-consistent relations between semantic entities and provides ground-truth labels of objects and scenes.

There are also works studying visual navigation under natural language guidance, including instruction following~\cite{chaplot2018gated,misra2017mapping,fried2018speaker,wang2018look,anderson2018vision} and question answering~\cite{das2018neural,das2017embodied,anand2018blindfold}. These tasks require the agent to understand the natural language and reason accordingly in an interactive environment. In our semantic navigation task, the goal instruction is simplified to a single semantic concept and hence we focus more on the reasoning ability of navigation agents. 

In our work, the reasoning is performed on the relations over semantic concepts  from visual signals. Similar ideas of using semantic knowledge to enhance reasoning have been applied to image classification~\cite{marino2016more,wang2018zero}, segmentation~\cite{torralba2003context,zhu2015segdeepm}, situation recognition~\cite{li2017situation}, visual question answering~\cite{wu2016ask,chen2018iterative,vedantam2019probabilistic,johnson2017clevr,hu2017modeling}, image retrieval~\cite{johnson2015image} and relation detection~\cite{zhang2017visual}. 
Savinov et al.~\cite{savinov2018episodic} and Kuang et al.~\cite{fang2019smt} also consider extracting visual concepts dynamically by treating every received input frame as an individual concept and storing them in the memory. 
The most related work to ours is a concurrent one by Wei et al.~\cite{WeiYang_ICLR2019}, which considers visual navigation towards an object category and utilizes a knowledge graph as the prior knowledge. Wei et al.~\cite{WeiYang_ICLR2019} use a fixed graph to extract features for the target  as an extra input to the locomotion without graph updating or planning. While in our work, the relation graph is used in a Bayesian manner as a representation for the memory, which unites use of prior knowledge, updating and planning altogether.

From a reinforcement learning perspective, our work is related to the model-based approaches~\cite{doya2002multiple,ross2008model,zhang2018composable,riedmiller2018learning,kurutach2018learning}, in the sense that we model the environment layout via a relation graph and plan on it. Our work is also related to hierarchical reinforcement learning~\cite{sutton1999between,dayan1993feudal,kulkarni2016hierarchical}, where the controller (BRM) produces a high-level sub-goal for the sub-policy (locomotion) to pursue. Furthermore, BRM learns from multi-task training and its update operation fast adapts the prior relations to the test environment, which can be also viewed as a form of meta-learning~\cite{finn2017model,duan2016rl,mishra2017meta}.

\section{Method}
\hide{
We first introduce the problem definition in Sec.~\ref{sec:def} and the overall architecture of our BRM agent (Fig.~\ref{fig:model}). We explain the memory module and the policy module respectively in Sec.~\ref{sec:brm} and Sec.~\ref{sec:policy}. Summary and discussions are in Sec.~\ref{sec:algo}.
}

\subsection{Task Setup}\label{sec:def}
We consider the semantic visual navigation problem where an agent interacts with an environment with discrete time steps and navigates towards a semantic goal. In House3D, the semantic entities of interest are room types and we assume a fixed number of $K$ categories. In the beginning of an episode, the agent is given a semantic target $T\in\mathcal{T}=\{T_1,\ldots,T_K\}$ to navigate towards for a success. At each time step $t$, the agent receives a visual observation $s_t$ and produces an action $a_t\in\mathcal{A}$ conditioning on $s_t$ and $T$. 

We aim to learn a neural agent that can \emph{generalize} to unseen environments. Hence, we train the agent on a training set $\mathcal{E}_{\textrm{train}}$, where the ground-truth labels are assumed, and validate on $\mathcal{E}_{\textrm{valid}}$. Evaluation is performed on another separate set of environments $\mathcal{E}_{\textrm{test}}$. At test time, the agent only access to the visual signal $s_t$ without any pre-exploration experiences of the test environment.

\subsection{Method Overview}\label{sec:overview}
The overall architecture of a BRM agent is shown in Fig.~\ref{fig:model}, which has two modules, the Bayesian Relational Memory (BRM) as well as an LSTM locomotor policy for control (Fig.~\ref{fig:model} left). Particularly, the key component of a BRM agent is a probabilistic relation graph (Fig.~\ref{fig:model} right), where each node corresponds to a particular semantic target $T_i$. For semantic target $T_i$ and $T_j$, the edge between them denotes the ``close-by'' relation and the probability of that edge implies how likely $T_i$ and $T_j$ are close to each other in the current environment.

At a high level, the locomotion is a semantic-goal-conditioned policy which takes in both the visual input $s_t$ and the sub-goal $g\in\mathcal{T}$ produced by the BRM module to produce actions towards $g$. The BRM module takes in the visual observation $s_t$ at each time step, extracts the semantic information via a CNN detector and store them in a replay buffer. We periodically update the posterior probability of each edge in the relation graph and re-plan to produce a new sub-goal. In our work, the graph is updated every fixed number of $N$ steps. Notably, we do not assume existences of all concepts --- in case of a missing concept in particular environment, the posterior of its associated edges will approach zero as more experiences gained in an episode. 

\begin{figure*}[bt!]
    \centering
    \includegraphics[width=0.49\textwidth]{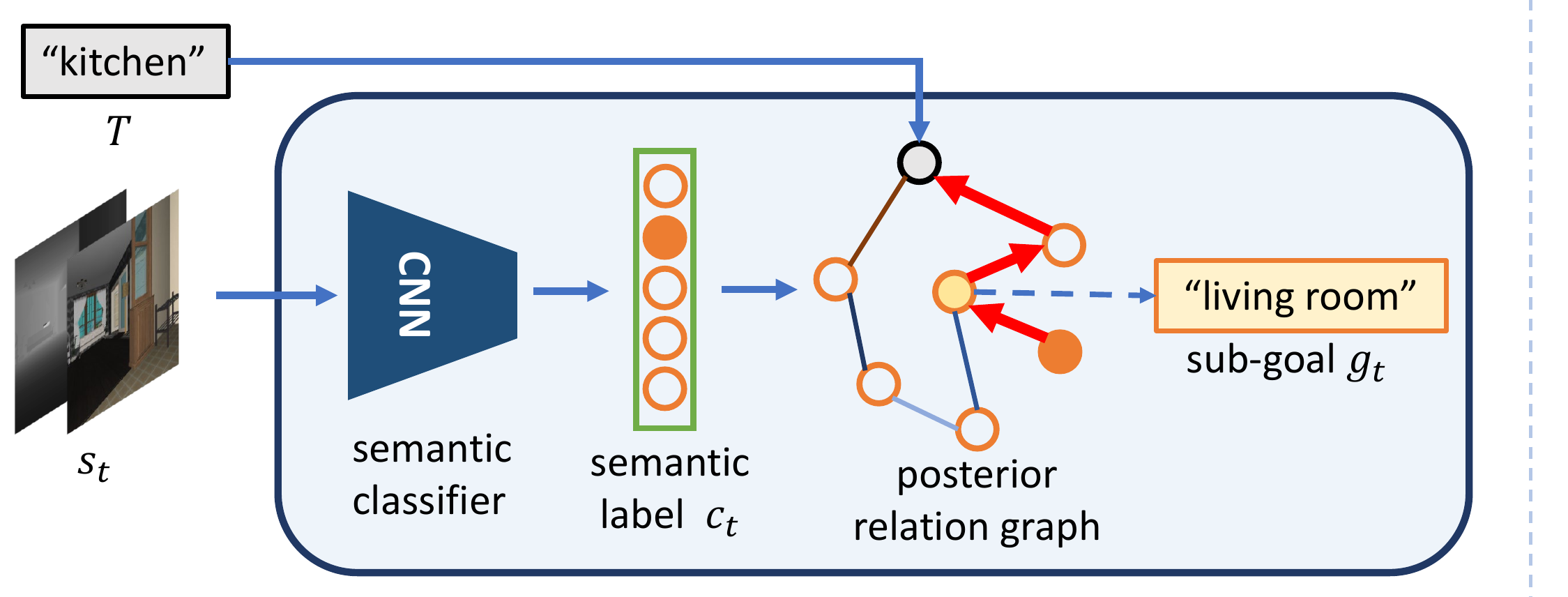}
    \includegraphics[width=0.49\textwidth]{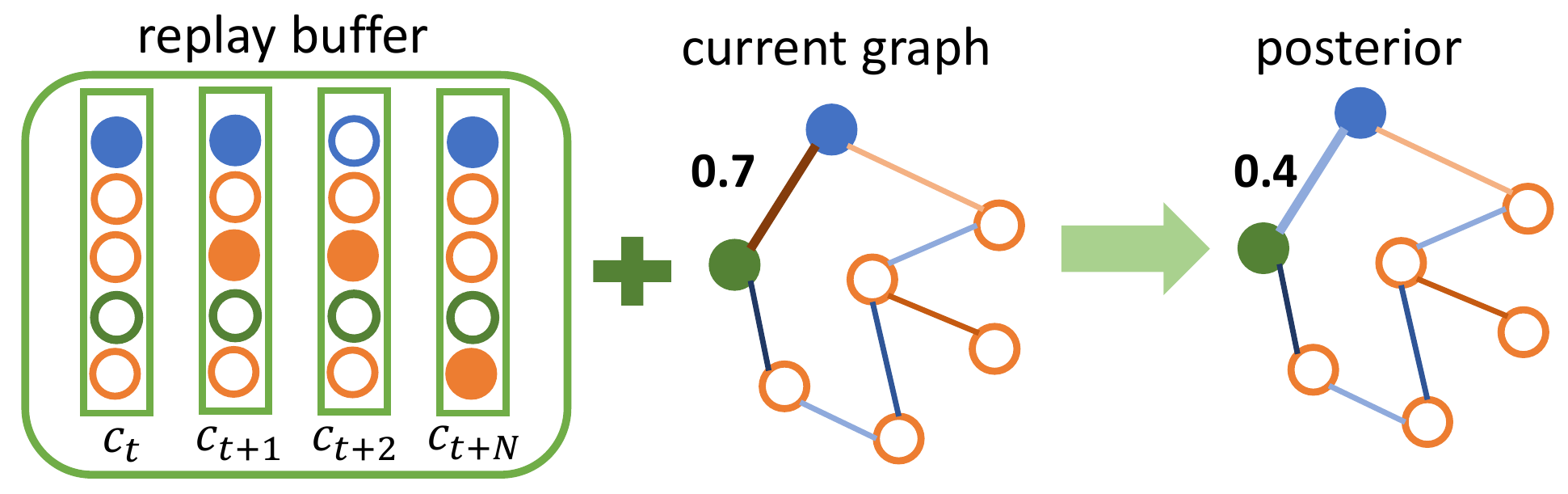}
    
    \caption{Overview of the planning (Left) and the update operation (Right) with BRM.}
    \label{fig:plan_update}
    \vspace{-2mm}
\end{figure*}

\subsection{Bayesian Relational Memory} \label{sec:brm}

The BRM module consists of two parts, a semantic classifier and the most important component, a probabilistic relation graph over semantic concepts, in the form of a probabilistic graphical model allowing efficient planning and posterior updates. 

Intuitively, at each time step, the agent detects the surrounding room type and maintains the probability of whether two room types $T_i$ and $T_j$ are ``nearby'' in the current environment. If the agent starts from room $T_i$ and reaches another room $T_j$ within a few steps, the probability of ``nearby'' relation between $T_i$ and $T_j$ should be increased\footnote{In perfect noiseless setting, the relation should with probability 1.}; otherwise the probability should be decreased. Periodically, the graph is updated and the agent finds the most likely path from the current room towards the target as a navigation guidance.

We introduce these two components in details as well as how to update and plan with BRM as follows.

%%%%%%%%%%%%%%%%%%%%%%%%%%%%%%%%%%%%%%%%%%%%%%%%%%%%%%%%

\noindent\textbf{Semantic classifier: }
The semantic classifier detects the room type label $c_t$ for the agent's surrounding region. It can be trained by supervised learning on $\mathcal{E}_\textrm{train}$.
Note that for robust room type classification, only using the first-person view image may not be enough. For example, the agent in a bedroom may face towards a wall, where the first-person image is not informative at all for the classifier, but a bed might be just behind.
So we take the panoramic view as the classifier input, which consists of 4 images, $s_t^{1},\ldots,s_t^{4}$ with different view angles. We use a 10-layer CNN with batch normalization to extract features $f(s_t^i)$ for each $s_t^i$ and compute the attention weights over these visual features by $l_i=f(s_o^i)W_1^TW_2\left[f(s_o^1),\ldots,f(s_o^4)\right]$ $a_i=\textrm{softmax}(l_i)$ with parameters $W_1,W_2$. Then the weighted average of these four features $\sum_i a_i f(s_t^i)$ is used for the final \texttt{sigmoid} predictions for each semantic concept $T_i$\footnote{It is a multi-label classification setting. Imagine an open kitchen with both cooking facilities and dining tables could have two labels.}. This results in a $K$-dimensional binary vector $c_t\in\{0,1\}^K$ at time $t$.

%%%%%%%%%%%%%%%%%%%%%%%%%%%%%%%%%%%%%%

\noindent\textbf{Probabilistic relation graph: }
We represent the probabilistic graph in the form of a graphical model $P(z,y;\psi)$ with latent variable $z$, observation variable $y$ and parameter $\psi$.

Since we have $K$ semantic concepts, there are $K$\footnote{In fact we have $K+1$ nodes. For clarity purpose, we use $K$ here and explain the details of the extra node later in Sec.~\ref{sec:algo}.} nodes and $K(K-1)/2$ relations (edges) in the graph. Each relation is probabilistic (i.e., it may exist or not with probability) and before entering a particular environment, we only hold a prior belief of that relation. Formally, for the relation between $T_i$ and $T_j$, we adopt a Bernoulli variable $z_{i,j}$ defined by $z_{i,j}\sim\mathrm{Bernoulli}(\psi^{\mathrm{prior}}_{i,j})$, where parameter $\psi_{i,j}^\mathrm{prior}$  denotes the prior belief of $z_{i,j}$ existing.
During exploration, the agent can noisily observe $z_{i,j}$ and use the noisy observations to estimate the true value of $z_{i,j}$. We define the noisy observation model $P(y_{i,j}|z_{i,j})$ by
\begin{equation}
y_{i,j} \sim \left\{
\begin{array}{ll}
\textrm{Bernoulli}(\psi^\mathrm{obs}_{i,j,0}) & \textrm{ if } z_{i,j}=0\\
\textrm{Bernoulli}(1 - \psi^\mathrm{obs}_{i,j,1}) & \textrm{ if } z_{i,j}=1
\end{array},
\right.
\label{eq:observ}
\end{equation}
where $\psi^\mathrm{obs}$ is another parameter to learn.
At each time step, the agent holds an overall posterior belief $P(z|\mathcal{Y})$ of relation existences within the current environment, based on its experiences $\mathcal{Y}$, namely the samples of variable $y$.

%%%%%%%%%%%%%%%%%%%%%%%%%%%%%%%%%%%%%%%%%%%%%%%%%%

\noindent\textbf{Posterior update and planning:}
A visualization of the procedures is shown in Fig.~\ref{fig:plan_update}. We assume the agent explores the current environment for a \emph{short} horizon of $N$ steps and stores the recent semantic signals ${c_t},\ldots,c_{t+N}$ in the replay buffer. Then we compute the bit-OR operation over these binary vectors 
$
B={c_t}\texttt{ OR }\ldots \texttt{ OR }{c_{t+N}}.
$
$B$ represents all the visited regions within a short ($N$-step) exploration period. When two targets appear concurrently in a short trajectory, they are assumed to be ``close-by''.
For $T_i$ and $T_j$ with $B(T_i)=B(T_j)=1$, $T_i$ and $T_j$ should be nearby in the current environment, namely a sample of $y_{i,j}=1$;  otherwise for $B(T_i)\ne B(T_j)$, we get a sample of $y_{i,j}=0$. With all the history samples of $y$ as $\mathcal{Y}$, we can perform posterior inference, i.e., compute posterior Bernoulli distribution $P(z_{i,j}|\mathcal{Y}_{i,j})$, for each $z_{i,j}$ by the Bayes rule.

Let $\hat{z}_{i,j}=P(z_{i,j}|\mathcal{Y}_{i,j})$ denote the posterior probability of relation over $T_i$ and $T_j$.
Given the current beliefs $\hat{z}$, the semantic signal $c_t$ and the target $T$, we search for an optimal plan $\tau^* = \{\tau_0, \tau_1, \ldots,\tau_{m-1}, \tau_m\}$ over the graph, where $\tau_i\in\{1\ldots K\}$ denotes an index of concepts, so that the joint belief along the path from some current position to the goal is maximized:
$\label{eq:plan}
    \tau^\star = \arg\max_\tau c_t(T_{\tau_0})\prod_{i=1}^m  \hat z_{\tau_{i-1}, \tau_{i}}.
$

After obtaining $\tau^\star$, we execute the locomotion policy for sub-goal $g_t=T_{\tau^\star_1}$, and then periodically update the graph, clear the replay buffer and re-plan every $N$ steps.

%%%%%%%%%%%%%%%%%%%%%%%%%%%%%%%%%%%%%%%%%%%%%%%%%%

\begin{figure}[bt!]
    \centering
    \includegraphics[width=0.45\textwidth]{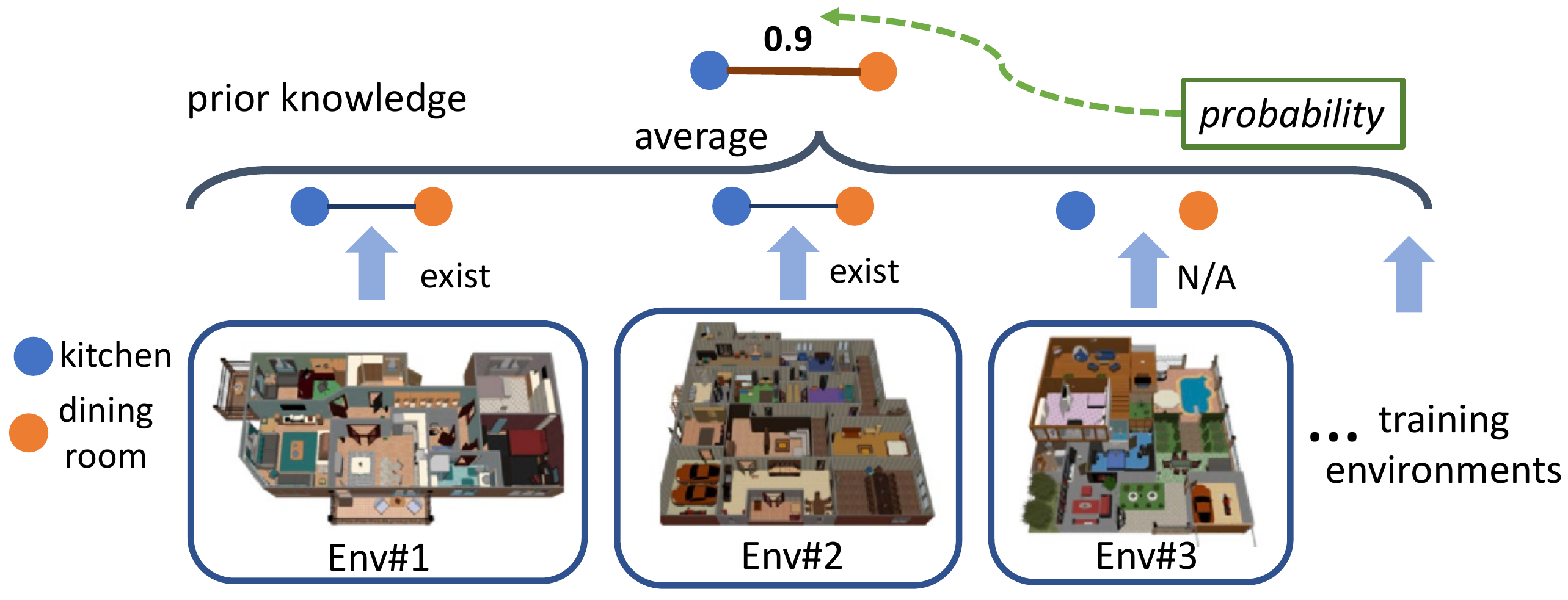}
    
    \caption{Learning the prior probability of a particular relation by analyzing the existence of that relation in each training environment and then summarizing the overall statistics.}
    \label{fig:prior}
    \vspace{-2mm}
\end{figure}

\begin{figure*}[bt]
\centering
\includegraphics[width=0.31\textwidth]{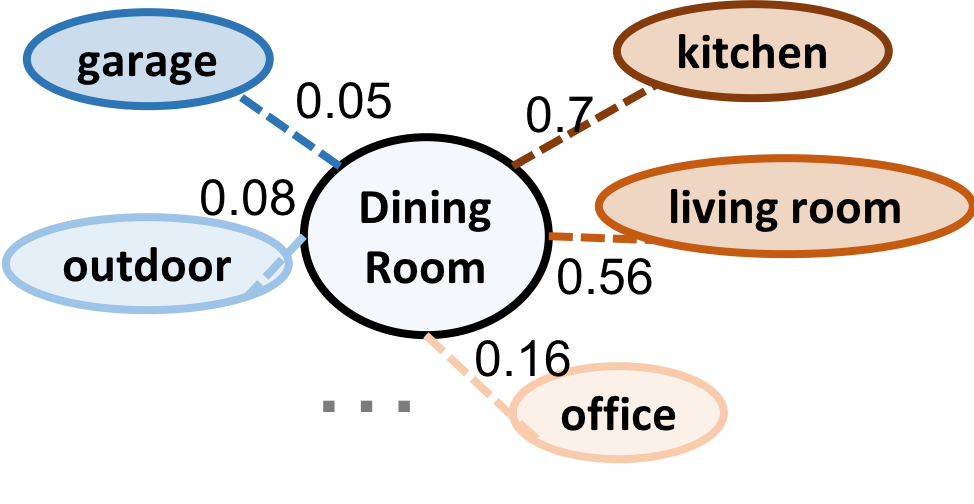}
\includegraphics[width=0.31\textwidth]{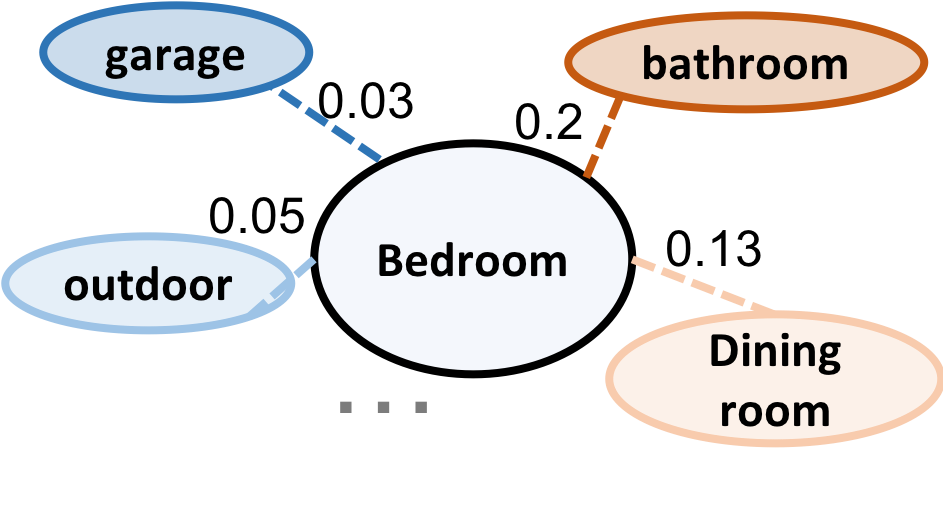}
\includegraphics[width=0.31\textwidth]{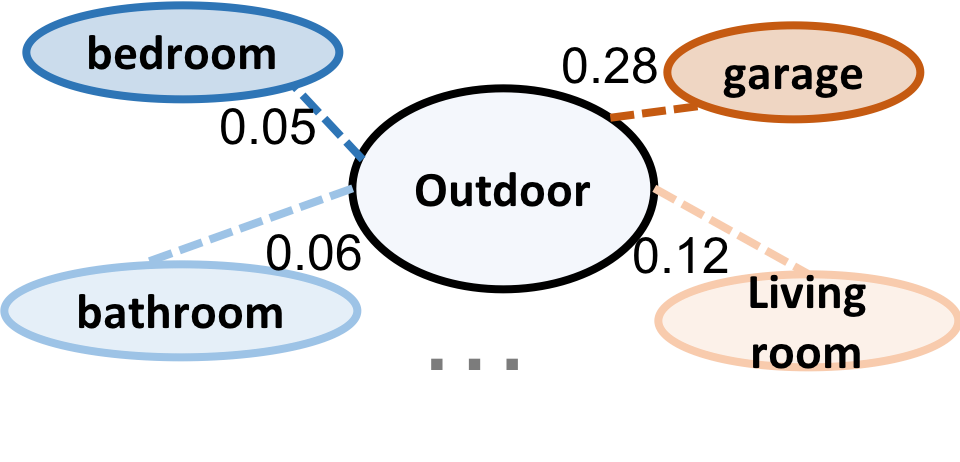}
\caption{The learned prior of the relations, including the most (red) and least (blue) likely nearby rooms for dining room (Left), bedroom (Mid.) and outdoor (Right), with numbers denoting $\psi^{\textrm{prior}}$, i.e., the prior probability of two room types are nearby. }
\label{fig:graph}
\vspace{-2mm}
\end{figure*}

\noindent\textbf{Learning the probabilistic graph: }
The parameter $\psi$ has two parts: $\psi^{\mathrm{prior}}$ for the prior of $z$ and $\psi^{\mathrm{obs}}$ for the noisy observation $y$. 

For the prior parameter $\psi^{\mathrm{prior}}$, we learn  from $\mathcal{E}_{\textrm{train}}$ with the ground truth labels. A visualization is shown in Fig.~\ref{fig:prior}. For each pair of room types $T_i$ and $T_j$, we enumerate all training environments and run \emph{random explorations} from some location of room type $T_i$. If eventually the agent reaches somewhere of room type $T_j$, we consider $T_i$ and $T_j$ are nearby and therefore a \emph{positive} sample $z_{i,j}=1$; otherwise a \emph{negative} sample $z_{i,j}=0$. 
Suppose $\mathcal{Z}$ denotes all the samples we obtained for $z$. We run maximum likelihood estimate for $\psi^{\textrm{prior}}$ by maximizing
$
L_{\textrm{MLE}}(\psi^{\textrm{prior}})=P(\mathcal{Z}|\psi^{\textrm{prior}})
$.
We can choose any exploration strategy such that for any ``close-by'' targets, they appear together in a short exploration trajectory more frequently. Random exploration is the most lightweight option among all.
%, the estimated probability $P(z_{i,j}=1)=\psi_{i,j}^{\textrm{prior}}$ should be higher.

The noisy observation parameter $\psi^{\mathrm{obs}}$ is related to the performance of the locomotion policy $\mu(\theta)$: if $\mu(\theta)$ has a higher navigation success rate, $\psi^{\mathrm{obs}}$ should be smaller (i.e., low noise level); when $\mu(\theta)$ is poor, $\psi^{\mathrm{obs}}$ should be larger (cf.~Eq.~\eqref{eq:observ}).  Unfortunately, there is no direct learning supervision for $\psi^{\mathrm{obs}}$. However, we can evaluate the ``goodness'' of a particular value of $\psi^{\mathrm{obs}}$ by evaluating the success rate of the overall BRM agent on $\mathcal{E}_{\textrm{valid}}$. Therefore, we can simply run grid search to derive the best parameter.

\hide{
We proceed to optimize the accumulative reward 
$
    L_{\textrm{valid}}(\psi^{\mathrm{obs}})=\mathbb{E}_{E\in\mathcal{E}_{\textrm{valid}}}\left[R(\mu(\theta),M(\psi); E)\right],
$ with BRM $M(\psi)$ to produce sub-goals.
Analytically optimizing $L_{\textrm{valid}}$ is hard. Instead, we apply grid search in practice to find the optimal $\psi^{\mathrm{obs}}$. 
}

\subsection{The Goal Conditioned Policy}\label{sec:policy}

We learn an LSTM locomotion policy $\mu(s_t,g;\theta)$ parameterized by $\theta$, which conditions on observation $s_t$ and navigates towards goal $g$.
Following Wu et al.~\cite{wu2018building}, we learn $\mu(s_t,g;\theta)$ by formulating the task as a reinforcement learning problem with shaped reward: when the agent moves towards target room $g$, it receives a positive reward proportional to the distance decrements; if the agent moves apart or hits an obstacle, a penalty is presented. A success reward of 10 and a time penalty of 0.1 are also assigned. We optimize the policy on $\mathcal{E}_{\textrm{train}}$ via the actor-critic method~\cite{mnih2016asynchronous} with a curriculum learning paradigm by periodically increasing the maximum spawn distance to the target $g$. 
Additionally, thanks to a limited set of $K$ targets, we adopt a behavior approach~\cite{chen2019behavioral} for improved performances: we train a separate policy $\mu_i(s_t;\theta_i)$ for each semantic target $T_i$ and when given the sub-goal $g$ from the BRM module, we directly execute its corresponding behavior network. We empirically observe it performs better than the original gated-attention policy in Wu et al.\cite{wu2018building}. Such an architecture is also common technique in other domains such as RL~\cite{oh2015action,finn2016unsupervised} and robotics~\cite{finn2017deep}.

\subsection{Implementation Details} \label{sec:algo}

\hide{
{
\begin{algorithm}[tb]
	\caption{Learning the BRM agent}\label{alg:leaps}
	\begin{algorithmic}[1]
		\REQUIRE $\mathcal{E}_{\textrm{train}}$, $\mathcal{E}_{\textrm{valid}}$, semantic concepts $T_{1\ldots K}$, 
		\STATE \textrm{// Step\#1: learn locomotion policy} %learning the sub-policy $\mu(s_t,g;\theta)$}
		%\STATE $\theta^\star = \arg\max_{\theta}\mathbb{E}_{T_i,E(c)\sim\mathcal{E}_{\textrm{train}}}\left[R(\mu(T_i,\theta);T_i,E(c))\right]$
		\STATE optimize $\mu(s_t,g;\theta)$ by actor critic method
		\STATE \textrm{// Part\#2: learning the BRM module}
		\STATE \textrm{learn the semantic classifier on $\mathcal{E}_{\textrm{train}}$}
		\FOR{$E(c)\in \mathcal{E}_{\textrm{train}}$}
		\FOR{$1\le T_i,T_j\le K$}
		\STATE spawn the agent to a room of type $T_i$
		\STATE random exploration and collect data into $\mathcal{Z}$
		\STATE repeat step 6 until enough data
		\ENDFOR
		\ENDFOR
		\STATE $\psi^{\textrm{prior}}=\arg\max_{\psi} P(\mathcal{Z}|\psi)$
		\STATE $\psi^{\textrm{obs}}=\arg\max_{\psi} \mathbb{E}_{E(c)
		\sim\mathcal{E}_{\textrm{valid}}}\left[R(\mu(\theta^\star,M(\psi);E(c)))\right]$
	\end{algorithmic}
\end{algorithm}

}
}

\hide{
The overall procedure for learning our BRM agent, i.e., a hierarchical navigation agent with BRM, is summarized in Algo.~\ref{alg:leaps}. We also introduce some crucial details below.}

We introduce those crucial details below and defer the remaining to appendix.

\noindent\textbf{Nodes in the relation graph: }
In the previous content, we assume $K$ pre-selected semantic concepts as graph nodes. However, it is not rare to reach some situation that cannot be categorized into any existing semantic category. In practice, we treat $c_t=\mathbf{0}$ as a special ``\emph{unknown}'' concept. Hence, the BRM module actually contains $K+1$ nodes. This is conceptually similar to natural language processing: a semantic concept is a word; the set of all concepts can be viewed as the dictionary; and $c_t=\mathbf{0}$ corresponds to the special ``out-of-vocabulary'' token. 

\noindent\textbf{Smooth temporal classification: }
Although the semantic classifier achieves high accuracy on validation data, the predictions may not be temporally consistent, which brings extra noise to the BRM module.
For temporally smooth prediction at test time, we set a restricted threshold over the sigmoid output and apply a filtering process on top of that: the actual prediction label for room type $T_i$ will be 1 only if the sigmoid output remains at least 0.9 confidence score for consecutively 3 time steps.

\noindent\textbf{Graph learning: }
For learning $\psi^{\textrm{prior}}$, we run a random exploration of 300 steps and collect 50 samples for each $z_{i,j}$ per training environment. Also, learning $\psi^{\textrm{prior}}$ does not depend on the locomotion  $\mu(s_t,g;\theta)$ and can be reused even with different control policies. $\psi^{\textrm{obs}}$ depends on the locomotion so it must be learned after $\mu(s_t,g;\theta^\star)$ is obtained.

\section{Experiments}
\label{sec:expr}
We experiment on the House3D environment and proceed to answer the following two questions: \textbf{(1)}  Does the BRM agent captures the underlying semantic relations and behave as we expected? \textbf{(2)} Does the BRM agent generalize better in test environments than the baseline methods? 

We first introduce evaluation preliminaries and baseline methods in Sec.~\ref{sec:prelim}, \ref{sec:baseline}. Then we answer the first question qualitatively in Sec.~\ref{sec:expr_qual}. In Sec.~\ref{sec:expr_results}, \ref{sec:expr_term}, we quantitatively show that our BRM agents generally achieve higher test success rates (i.e., better generalization) and spend fewer steps to reach the targets (i.e., more efficient) than all the baselines. We choose a fixed $N=10$ for all the BRM agents. Ablation study on the design choices of BRM is presented in Sec.~\ref{sec:ablation}. More details can be found in appendix.

%%%%%%%%%%%%%%%%%%%%%%%%%%%%%

\begin{figure}[bt]
	\centering
	\includegraphics[width=0.4\textwidth]{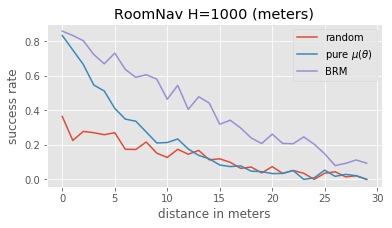}
	\caption{Qualitative comparison between BRM (purple), basic DRL policy (pure $\mu(\theta)$, blue) and random policy (red) with horizon $H=1000$ (Sec.~\ref{sec:expr_qual}). Y-axis is the test success rate and x-axis is the distance in meters to the target. When targets become farther way from the starting point, the success rate of BRM stays high while the basic DRL policy quickly degenerates to random. }
	\label{fig:roomnav}
	\vspace{-2mm}
\end{figure}

\subsection{Preliminaries}\label{sec:prelim}

We consider the RoomNav task on the House3D environment~\cite{wu2018building} where $K=8$ room types are selected as the semantic goals, including ``kitchen'', ``living room'', ``dining room'', ``bedroom'', ``bathroom'', ``office'', ``garage'' and ``outdoor''. The House3D environment provides a success check for whether the agent has reached a specific room target or not while we also experimented on the setting of the agent predicting termination on its own (Sec.~\ref{sec:expr_term}). House3D provides a training set of 200 houses, a test set of 50 houses and a validation set of 20 houses. At training time, all the approaches adopt the ground-truth semantic labels regardless of the semantic classifier.

We evaluate the generalization performances of different approaches with two metrics, \emph{success rate} and \emph{Success weighted by Path Length (SPL)}, under different horizons. SPL, proposed by Anderson et al.\citet{anderson2018evaluation}, is a function of both success rate and episode length defined by $\frac{1}{C}\sum_i S_i\frac{L_i}{\textrm{max}(L_i,P_i)}$, where $C$ is total episodes evaluated, $S_i$ indicates whether the episode is success or not, $L_i$ is the ground truth shortest path distance in the episode, $P_i$ is the number of steps the agent actually took. SPL is upper-bounded by success rate and assigns more credits to agents accomplishing their tasks faster.

\subsection{Baseline Methods}\label{sec:baseline}
\noindent\textbf{Random policy: } The agent samples a random action per step, denoted by ``random''.

\noindent\textbf{Pure DRL agent: } This LSTM agent does not have the BRM module and directly executes the policy $\mu(s_t,T;\theta)$ throughout the entire episode, denoted by ``pure $\mu(\theta)$''. This is in fact the pure locomotion module. As discussed in Sec.~\ref{sec:policy}, this is also an improved version of the original policy proposed by Wu et al.~\cite{wu2018building}.

\noindent\textbf{Semantic augmented agent: }
Comparing to the pure DRL agent, our BRM agents utilizes an extra semantic signals $c_t$ provided by the semantic classifier in addition to the visual input $s_t$. Hence, we consider a semantic-aware locomotion baseline $\mu_S(\theta_s)$, which is another LSTM DRL agent that takes both $s_t$ and $c_t$ as input (denoted by ``aug.$\mu_S(\theta_s)$'').

\noindent\textbf{HRL agent with an RNN controller: } 
From a DRL perspective, our BRM agent is a hierarchical reinforcement learning (HRL) agent with the BRM module as a high-level controller producing sub-goals and the locomotion module as a low-level policy for control. Note that update and planning on BRM only depend on (1) the current semantic signal $c_t$, (2) the target $T$, and (3) the accumulative bit-\texttt{OR} feature $B$ (see Sec.~\ref{sec:brm}). Hence, we adopt the same locomotion $\mu(s_t,g;\theta)$ used by our BRM agent, and train an LSTM controller with 50 hidden units on $\mathcal{E}_{\textrm{train}}$ that takes all the necessary semantic information and produces a sub-target every $N$ steps as well. %Training details are in Appendix~\ref{app:baseline}.
The only difference between our BRM agent and this HRL agent is the \emph{representation} of the controller (memory) module. The LSTM controller has access to exactly the same semantic information as BRM and uses a much more complex and generic neural model instead of a relation graph. Thus, we expect it to be a strong baseline and perform competitively to our BRM agent.

%%%%%%%%%%%%%%%%%%%%%%%%%%%%%%%%

\begin{figure*}[bt!]
    \centering
    \includegraphics[width=0.95\linewidth]{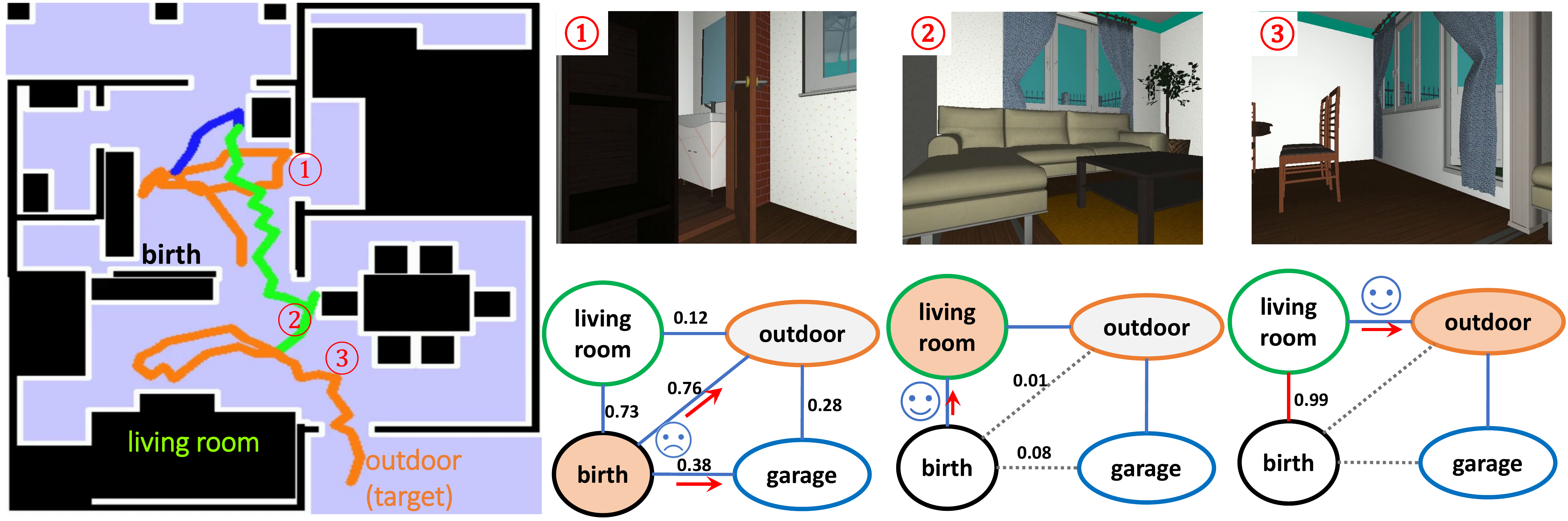}
    \caption{Example of a successful trajectory. The agent is spawned inside the house, targeting ``outdoor''. \textbf{Left}: the 2D top-down map with goal-conditioned trajectories (``outdoor'' -- orange; ``garage'' -- blue; ``living room'' -- green); \textbf{Right, 1st row}: RGB visual image; \textbf{Right, 2nd row}: the posterior of the semantic graph and the proposed sub-goals (red arrow). Initially, the agent starts by executing the locomotion for  "outdoor" and then "garage" according to the prior knowledge (\textbf{1st graph}), but both fail (top orange and blue trajectories in the map). After updating its belief that garage and outdoor are not nearby (grey edges in the \textbf{2nd graph}), it then executes locomotion  for "living room" with success (red arrow in the \textbf{2nd graph}, green trajectory). Finally, it executes sub-policy for``outdoor'' again, explores the living room and reaches the goal (\textbf{3rd graph}, bottom orange trajectory).}
    \label{fig:casestudy_outdoor}
    \vspace{-2mm}
\end{figure*}

\subsection{Qualitative Analysis and Case Study}
\label{sec:expr_qual}
In this section, we qualitatively illustrate that our BRM agent is able to learn reasonable semantic relations and behave in an interpretable manner. 

\noindent\textbf{Prior knowledge:} Fig.~\ref{fig:graph} visualizes $P(z|\psi^{\mathrm{prior}})$, the learned prior probability of relations, for 3 room types with their most and least likely nearby (connected) rooms. Darker red means more likely while darker blue implies less likely. The captured knowledge is indeed reasonable: bathroom is likely to connect to a bedroom; kitchen is often near a dining room while garage is typically outdoor.

\noindent\textbf{Effectiveness of planning: }
The BRM agent can effectively decompose a long-term goal into a sequence of easier sub-goals via graph planning.
Fig.~\ref{fig:roomnav} visualizes the test success rates of the BRM agent (BRM), random policy (``random'') and the pure locomotion policy (``pure $\mu(\theta)$'') for increasingly further targets over a fixed set of 5689 randomly generated test tasks. The x-axis is the shortest distance to the target in meters\footnote{Typically one meter in shortest distance requires 3 to 4 actions.} As expected, when the target becomes more distant, all methods have lower success rates. However, as opposed to the pure locomotion policy, which quickly degenerates to random as the distance increases, the BRM agent remains a much higher success rate in general.

\noindent\textbf{Case study:} Fig.~\ref{fig:casestudy_outdoor} shows a successful trajectory by the BRM agent, where the final target is to get out of the house. We visualize the progression of the episode, describe the plans and show the updated graph during exploration. Note that the final goal is invisible to the agent initially (frame \textcircled{1}) but the agent is able to plan, effectively explore the house (e.g., without ever entering the bottom-right dining room region), and eventually reach the target.

\subsection{Quantitative Generalization Performances}
\label{sec:expr_results}
We evaluate the generalization performances of different approaches on $\mathcal{E}_{\textrm{test}}$ with horizons $H=300$ and $H=1000$. We set $N=10$, i.e., memory updated every 10 steps. Tab.~\ref{fig:eval} reports both success rates~(\%, percent) and SPL values~(\textperthousand, per mile) over 5689 fixed test tasks. In addition to the overall performances, we also report the results under different \emph{planning distances}, i.e., the shortest sequence of sub-goals on the ground-truth relation graph. 
For an accurate measurement, we ensure that there are at least 500 test tasks for each planning distance.

%BRM generally outperforms all the baselines. %Further targets are typically more challenging. 

Our BRM agent has the highest average success rates as well as the best SPL values in \emph{all} the cases. Notably, the margin in SPL is much more significant than that in pure success rate. 
More importantly, as the horizon increases, i.e., the larger number of planning computations $H/N$ allowed, the overall performance margin (rightmost column) of BRM over the best remaining baseline strictly increases, thanks to the effectiveness of planning.

\hide{
Note that BRM improves little in success rates does not improve much over the baselines for targets of plan-step 1, which do not require any semantic planning. Thus similar performances are as expected. We also notice that in shorter horizons ($H\le 500$), for plan-steps equal to 4, BRM agents have the highest success rate but relatively lower SPL. This is because that BRM utilizes a fixed update frequency for memory update, which may potentially increase the episode length for goals requiring more planning computations. When the horizon is long, i.e., allowing enough BRM updates and planning, our BRM agents significantly outperform all the baselines in SPL. It can be helpful if BRM learns to update the graph adaptively. We leave this to future work.
}

\begin{table*}[bt]
	\centering
	\begin{tabular}{c|c|c|c|c|c|c}
		\hline
		opt. plan-steps&1&2&3&4&5&overall\\
		\hline
		\hline
		avg. oracle steps & 12.27&42.53&61.09&72.47&63.74&46.86\\
		\hline
		\hline
		\multicolumn{7}{c}{Horizon $H=300$}\\
		\hline
		\hline
random& 20.5 / 15.9& 6.9 / 16.7& 3.8 / 10.7& 1.6 / 4.2& 3.0 / 8.8& 7.2 / 13.6\\
\hline
pure $\mu(\theta)$& 49.4 / 47.6& 11.8 / 27.6& 2.0 / 4.8& 2.6 / 10.8& 4.2 / 13.2& 13.1 / 22.9\\
\hline
aug.$\mu_S(\theta)$& 47.8 / 45.3& 11.4 / 23.1& 3.0 / 7.8& 3.4 / 8.1& 4.4 / 11.2& 13.0 / 20.5\\
\hline
RNN control.& 55.0 / 49.8& 20.0 / 40.8& 8.0 / 20.1& 5.2 / 15.2& 11.0 / 25.2& 19.9 / 34.2\\
\hline
BRM& \textbf{57.8} / \textbf{65.4}& \textbf{24.4} / \textbf{54.3}& \textbf{10.5} / \textbf{28.3}& \textbf{5.8} / \textbf{18.6}& \textbf{11.2} / \textbf{29.8}& \textbf{23.1} / \textbf{45.3}\\
\hline
		\hline
		\multicolumn{7}{c}{Horizon $H=1000$}\\
		\hline
		\hline
plan-dist&1&2&3&4&5&avg.\\
\hline
\hline
random& 24.3 / 17.6& 13.5 / 20.3& 9.1 / 14.3& 8.0 / 9.3& 7.0 / 11.5& 13.0 / 17.0\\
\hline
pure $\mu(\theta)$& 60.8 / 47.6& 23.3 / 27.6& 7.6 / 4.8& 8.2 / 10.8& 11.0 / 13.2& 22.5 / 22.9\\
\hline
aug.$\mu_S(\theta)$& 61.3 / 50.1& 23.0 / 26.2& 9.4 / 12.0& 5.8 / 9.6& 9.0 / 13.6& 22.4 / 23.8\\
\hline
RNN control.& 71.0 / 58.0& 39.6 / 51.3& 24.1 / 32.7& 16.6 / 25.6& 23.2 / 39.6& 37.0 / 45.2\\
\hline
BRM& \textbf{73.7} / \textbf{74.9}& \textbf{43.6} / \textbf{66.0}& \textbf{29.2} / \textbf{44.9}& \textbf{20.4} / \textbf{27.1}& \textbf{28.4} / \textbf{42.5}& \textbf{41.1} / \textbf{57.5}\\
\hline

	\end{tabular}
	\caption{Metrics of \textbf{\emph{Success Rate(\%) / SPL(\textperthousand, per mile)}} evaluating the generalization performances of BRM and all the baseline approaches (Sec.~\ref{sec:expr_results}). Here $N=10$.  ``\emph{plan-steps}'' denotes the shortest planning distance in the ground truth relation graph. ``oracle steps'' denotes the reference shortest steps required to reach the goal.  Our BRM agents have the highest success rates and the best SPL values in all the cases. More importantly, as the horizon increases, which allows more planning, BRM outperforms the baselines more. } 
	\label{fig:eval}
	\vspace{-1mm}
\end{table*}

\hide{
\begin{table*}[bt]
	\centering
	\begin{tabular}{c|c|c|c|c|c}
		\hline
		\hline
		\multicolumn{5}{c}{Horizon $H=1000$ with Terminate Action}\\
		\hline
		\hline
		random&pure $\mu(\theta)$&aug.$\mu_S(\theta_s)$&RNN control. &BRM\\
		\hline
		2.1 / 1.3&9.6 / 9.7&9.0 / 9.4&15.9 / 17.6&\textbf{18.7} / \textbf{24.2}\\
		\hline
		
	\end{tabular}
	\caption{Metrics of \textbf{\emph{Success Rate(\%) / SPL(\textperthousand, per mile)}} with \textbf{terminate action} evaluating the generalization performances of BRM and baseline agents with horizon $H=1000$ and $N=10$. Our BRM agent still achieves the best performance under both metrics. } 
	\label{fig:term}
\end{table*}
}

\subsection{Ablation Study}\label{sec:ablation}

In this section, we show the necessity of all the BRM components and the direction for future improvement.

\noindent\textbf{Benefits of Learned Prior:} In BRM, the prior $P(z|\psi^{\textrm{prior}})$ is learned from training houses. Tab.~\ref{fig:uniprior} evaluates BRM with an \emph{uninformative prior} (``unif.''), i.e. $\psi^{\textrm{prior}}_{i,j}=0.5$. Generally, the learned prior leads to better success rates and SPL values. Notably, when horizon becomes longer, the gap becomes much smaller, since the graph will converge to the true posterior with more memory updates.

%Learned prior leads to better SPLs (fewer steps) and better success rates under shorter horizon.
\begin{table}[bt]
%\vspace{-6mm}
	\centering
	\begin{tabular}{c|c|c|c}
		\hline
		\footnotesize{BRM ($H$=300)} & \footnotesize{unif. ($H$=300)}&\footnotesize{BRM ($H$=1k)} & \footnotesize{unif. ($H$=1k)}\\
		\hline
		\textbf{23.1} / \textbf{45.3}&20.9 / 39.4 &\textbf{41.1} / \textbf{57.5}& 40.4 / 56.6\\
		\hline
		
	\end{tabular}
	%\vspace{-2mm}
	\caption{\textbf{\emph{Success Rate(\%) / SPL(\textperthousand)}}: performances of BRM with learned and uninformative prior. $N=10$, $H=300, 1000$ (``1k''). } 
	\label{fig:uniprior}
	%\vspace{-1mm}
\end{table}

\noindent\textbf{Source of Error: } Our approach has two modules, a BRM module for planning and a locomotion module for control. We study the errors caused by each of these components by introducing (1) a hand-designed (imperfect) oracle locomotion, which automatically gets closer to nearby targets\hide{The oracle locomotion is designed as follows: (1) if the target within 3 meters, the agent succeeds with 85\% prob. or becomes 50\% closer; (2) if target is away, the agent randomly selects another target less than 10 meters away and moves towards it at least 5 meters.}, and (2) an optimal planner using House3D labels.  The evaluation results are show in Tab.~\ref{fig:error}, where the oracle locomotion drastically boosts the results while the BRM performance is close to the optimal planner. This indicates that the error is mainly from locomotion -- it is extremely challenging to learn a single neural navigator, which motivates our work to decompose a long-term task into sub-tasks. 
\begin{table}[bt]
	\centering
	%\vspace{-3mm}
	\begin{tabular}{c|c|c}
		%\hline
		%\hline
		%\multicolumn{3}{c}{Source of Error, $H=1000$, $N=30$}\\
		\hline
		%\hline
		Src. of Err.&LSTM locomotion & oracle locomotion\\
		\hline
		BRM& 41.1 / 57.5 &88.6 / N.A.\\
		\hline
		opt. plan&46.3 / 62.5&96.7 / N.A.\\
		\hline
		
	\end{tabular}
	\caption{\textbf{\emph{Success Rate(\%) / SPL(\textperthousand)}}: performances with an optimal planner and a hand-designed locomotion. $H$=1000, $N$=10.} 
	\label{fig:error}
	%\vspace{-1mm}
\end{table}

\noindent\textbf{Choice of $N$:} 
The oracle steps for reaching a nearby, i.e., 1-plan-step, target is around 12.27 (top in Tab.~\ref{fig:eval}), so we choose a slightly smaller value $N=10$ as the re-planning step size. We investigate other choices of $N$ in Tab.~\ref{fig:choice}. Larger $N$ results in more significant performance drops. Notably,  our BRM agent consistently outperforms RNN controller under different parameter choices.

\begin{table}[bt]
	\centering
	%\vspace{-5mm}
	\begin{tabular}{c|c|c|c}
		\hline
		Choice of $N$&$N=10$ &$N=30$&$N=50$\\
		\hline
		BRM& \textbf{41.1} / \textbf{57.5}&29.7 / 35.2&27.4 / 32.2\\
		\hline
		RNN control.&37.0 / 45.2 &28.2 / 27.7&26.5 / 26.7\\
		\hline
		
	\end{tabular}
	\caption{\textbf{\emph{Success Rate(\%) / SPL(\textperthousand)}}: performances with different choices of $N$ under horizon $H=1000$.} 
	\label{fig:choice}
	%\vspace{-1mm}
\end{table}

\begin{table}[bt]
	\centering
	%\hspace{-2mm}
	\small{
	\begin{tabular}{c|c|c|c|c}
		\hline
		\hline
		\multicolumn{5}{c}{Horizon $H=1000$ with Terminate Action}\\
		\hline
		\hline
		random&pure\,$\mu(\theta)$&aug.$\mu_S(\theta_s)$&RNN cont. &BRM\\
		\hline
		1.8\hspace{0.05em}/\hspace{0.05em}1.2&8.6\hspace{0.07em}/\hspace{0.07em}9.0&8.2\hspace{0.07em}/\hspace{0.07em}8.8&14.5\hspace{0.07em}/\hspace{0.07em}16.3&\textbf{17.3}\hspace{0.05em}/\hspace{0.05em}\textbf{23.0}\\
		\hline
		
	\end{tabular}
	}
	\caption{\textbf{\emph{Success Rate(\%) / SPL(\textperthousand)}} with \textbf{terminate action} evaluating the generalization performances of BRM and baseline agents with horizon $H=1000$ and $N=10$. Our BRM agent achieves the best performances under both metrics. } 
	\label{fig:term}
	%\vspace{-1mm}
\end{table}

\subsection{Evaluation with Terminate Action}\label{sec:expr_term}
In the previous studies, the success of an episode is determined by the House3D environment automatically. It is suggested by Anderson et al.~\citet{anderson2018evaluation} that a real-world agent should be aware of its goal and determine whether to stop by itself. In this section, we evaluate the BRM agent and all previous baselines under this setting: a success will be counted only if the agent terminates the episode correctly in a target room on its own. 

There are two ways to include the terminate action: (1) expand the action space with an extra stop action; (2) train a separate termination checker. We observe (2) leads to much better practical performances, which is also reported by Pathak et al.~ \citet{pathakICLR18zeroshot}. In our experiments, we simply use the semantic classifier as our termination checker.

The results are summarized in Tab.~\ref{fig:term}, where we use a long horizon $H=1000$ to allow the agents to have enough time to self-terminate. Similarly, BRM achieves the best performance in both success rate and SPL metric.

\subsection{Discussions}\label{sec:discuss}
\noindent\textbf{Success rate and SPL:} In Tab.~\ref{fig:eval}, the SPL values are typically much smaller than the success rates, namely BRM uses much more steps than the reference shortest path. This is not surprising due to the strong partial observability in the RoomNav task. As a concrete example in Fig.~\ref{fig:casestudy_outdoor}, the optimal path from birthplace (near \textcircled{1}) to the outdoor (near \textcircled{3}) is extremely short if we know the top-down view in advance. However, the outdoor region is out of the agent's sight (frame \textcircled{1}) and the agent has to \emph{explore} the nearby regions before it sees the door towards the outside (frame \textcircled{3}). 
Planning and updates on BRM helps guide the agent to explore the unknown house more effectively, which helps lead to higher SPL values. But overall the agent still suffers from the fundamental challenge of partial observability.

\noindent\textbf{Pre-selected concepts: } In this work, we focus on the memory representation and simply assume we know all the semantic concepts in advance. It is also possible to generalize to unseen concepts by leveraging the word embedding and knowledge graph from NLP community~\cite{WeiYang_ICLR2019}. It is also feasible to directly discover general semantic concepts from $\mathcal{E}_{\textrm{train}}$ via unsupervised learning by leveraging the rich semantic information (e.g., object categories, room types, etc) within the visual input. We leave this to our future work.

%Existing works have shown promising results for vision-and-language tasks via learning cross-modal embedding from joint vision and natural language input~\cite{niu2017hierarchical,frome2013devise,karpathy2015deep}.

\hide{
\textbf{Pre-selected concepts:} 
In this work, we only consider the simplest case with $K=8$ concepts but our framework can be extended  to further bridge the gap between semantic reasoning and continuous control by leveraging the massive data and knowledge from natural language domain. 

It is also feasible to directly discover semantic concepts from $\mathcal{E}_{\textrm{train}}$ via unsupervised learning. This is beyond the focus of this work and we leave it as future work.
}

\section{Conclusion}

In this work, we proposed a novel design of memory architecture, Bayesian Relation Memory (BRM), for the semantic navigation task. BRM uses a semantic classifier to extract semantic labels from visual input and builds a probabilistic relation graph over the semantic concepts, which allows representing prior reachability knowledge via the edge priors, fast test-time adaptation via edge posteriors and efficient planning via graph search. Our BRM navigation agent uses BRM to produce a sub-goal for the locomotion policy to reach. Experiment results show that the BRM representation is effective and crucial for a visual navigation agent to generalize better in unseen environments.

At a high-level, our approach is general and can be applied to other tasks with semantic context information or state abstractions available to build a graph over, such as robotics manipulations where semantic concepts can be abstract states of robot arms and object locations, or video games where we can plan on semantic signals such as the game status or current resources. In future work, it is also worthwhile to investigate how to extract relations and concepts directly from training environments automatically. 
\newpage
{\small
\bibliographystyle{ieee_fullname}
\bibliography{references}
}

\clearpage

\appendix

\begin{figure*}[bt!]
	\centering
	\includegraphics[width=0.45\textwidth]{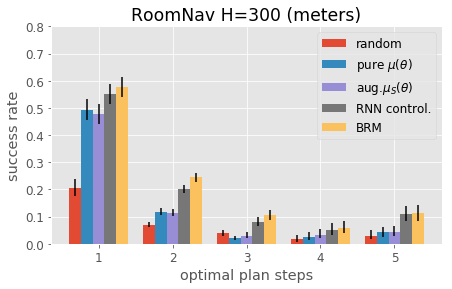}
	\includegraphics[width=0.45\textwidth]{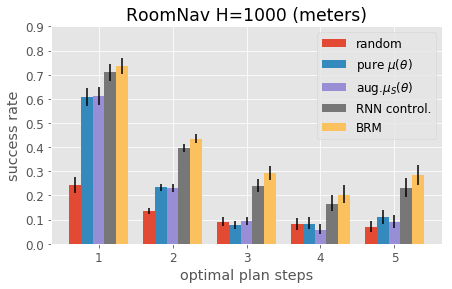}
	\caption{Comparing BRM with baselines in success rate with confidence interval. Approaches of interest include random policy (red), pure LSTM policy (blue), the semantic-aware LSTM policy (purple), the hierarchical policy (grey) and BRM (yellow). In all plots, the y-axis is success rate while the x-axis is the optimal planning distance. BRM outperforms all baselines and the gap becomes more significant when horizon increase, namely, more planning computations.\vspace{-2mm}}
	\label{fig:conf}
\end{figure*}

\section{Video Demo}
A video demo visualizing a successful navigation trajectory by BRM can be found at the following url:

\small{\url{https://drive.google.com/file/d/1vCFQZfFK1X6WJacrQID2kMQVzRD4WeTs/view?usp=sharing}}.

\section{Environment Details}\label{app:env}
In RoomNav the 8 targets are: kitchen, living room, dining room, bedroom, bathroom, office, garage and outdoor.
%For ObjectNav, the 15 object targets are: kitchen cabinet, sofa, chair, toilet, table, sink, wardrobe cabinet, bed, shelving, desk, television, household appliance, dresser, vehicle and pool. 
We inherit the success measure of ``see'' from \cite{wu2018building}: the agent needs to see some corresponding object for at least 450 pixels in the input frame and stay in the target area for at least 3 time steps.

Originally the House3D environment supports a set of 13 discrete actions. Here we reduce it to 9 actions: large forward, forward, left-forward, right-forward, large left rotate, large right rotate, left rotate, right rotate and stay still. More environment details can be found in the appendix of Wu et al.~\cite{wu2018building}. We also implemented a faster and customized variant of the House3D environment, which is available at \url{https://github.com/jxwuyi/House3D/tree/C++}.

\section{Evaluation Details}\label{app:eval}

We measure the success rate on $\mathcal{E}_{\textrm{test}}$ over 5689 test episodes, which consists of 5000 randomly generated configurations and 689 specialized for faraway targets to increase the confidence of measured success rates. These 689 episodes are generated such that for each plan-distance, there are at least 500 evaluation episodes. Each test episode has a fixed configuration for a fair comparison between different approaches, i.e., the agent will always start from the same location with the same target in that episode. Note that we always ensure that (1) the target is connected to the birthplace of the agent, and (2) the the birthplace of the agent is never within the target room. 
In addition to the detailed numbers in Table~\ref{fig:eval}, 
we visualize the success rates with \emph{confidence intervals} for BRM and baseline methods in Figure~\ref{fig:conf}. The confidence interval is obtained by fitting a binomial distribution.

\section{Ablation Study: the Semantic Detector}\label{sec:oracle}
In BRM, we use a CNN detector to extract the semantic signals at test time. Here we also evaluate the performances of all the approaches using the oracle signals from the House3D environment. The results are in Table~\ref{fig:oracle}, where we also include the BRM agent using CNN detector as a reference.  Generally, using both the ground truth signal and using the CNN detector yield comparable overall performances in both metrics of success rate and SPL. They all consistently outperform all the baseline methods, which indicates that the probabilistic relational graph is robust over the noise on semantic signals (the robustness if controlled by $\psi^{\textrm{obs}}$). One interesting observation is that there are many cases, using CNN detector produces better results than using the ground truth signals. We hypothesis that this is because the semantic labels in House3D is noisy and therefore a well-trained CNN detector will not be influenced by the noisy labels at test time. 
\begin{table*}[h]
    \centering
    \begin{tabular}{c|c|c|c|c|c|c}
\hline
plan-dist&1&2&3&4&5&overall\\
\hline
\hline
\multicolumn{7}{c}{Horizon $H=300$}\\
\hline
\hline
plan-dist&1&2&3&4&5&avg.\\
\hline
\hline
random& 20.5 / 15.9& 6.9 / 16.7& 3.8 / 10.7& 1.6 / 4.2& 3.0 / 8.8& 7.2 / 13.6\\
\hline
pure $\mu(\theta)$& 49.4 / 47.6& 11.8 / 27.6& 2.0 / 4.8& 2.6 / 10.8& 4.2 / 13.2& 13.1 / 22.9\\
\hline
aug.$\mu_S(\theta)$ (true)& 51.9 / \textbf{66.4}& 11.1 / 24.2& 3.3 / 7.8& 2.4 / 6.0& 3.0 / 8.7& 13.2 / 23.3\\
\hline
RNN control. (true)& 54.9 / 48.1& 20.2 / 37.7& 8.2 / 22.5& 5.6 / 13.8& 9.8 / 22.7& 20.0 / 32.6\\
\hline
BRM (true)& \textbf{58.8} / 60.7& \textbf{25.3} / \textbf{55.6}& 10.4 / 26.9& \textbf{7.6} / \textbf{22.2}& 9.2 / 23.4& \textbf{23.6} / 44.9\\
\hline
BRM (CNN)& 57.8 / 65.4& 24.4 / 54.3& \textbf{10.5} / \textbf{28.3}& 5.8 / 18.6& \textbf{11.2} / \textbf{29.8}& 23.1 / \textbf{45.3}\\
\hline
\hline
\multicolumn{7}{c}{Horizon $H=1000$}\\

\hline
\hline
plan-dist&1&2&3&4&5&avg.\\
\hline
\hline
random& 24.3 / 17.6& 13.5 / 20.3& 9.1 / 14.3& 8.0 / 9.3& 7.0 / 11.5& 13.0 / 17.0\\
\hline
pure $\mu(\theta)$& 60.8 / 47.6& 23.3 / 27.6& 7.6 / 4.8& 8.2 / 10.8& 11.0 / 13.2& 22.5 / 22.9\\
\hline
aug.$\mu_S(\theta)$ (true)& 62.4 / 61.3& 22.9 / 30.7& 8.9 / 14.3& 7.2 / 12.8& 9.0 / 11.4& 22.5 / 28.1\\
\hline
RNN control. (true)& 70.2 / 51.3& 40.8 / 48.6& 22.8 / 32.2& 16.4 / 23.4& 24.2 / 41.0& 37.4 / 42.9\\
\hline
BRM (true)& 70.3 / 61.8& \textbf{44.9} / \textbf{70.5}& \textbf{31.7} / \textbf{50.8}& 19.0 / \textbf{33.3}& 28.0 / 42.2& \textbf{41.7} / \textbf{59.8}\\
\hline
BRM (CNN)& \textbf{73.7} / \textbf{74.9}& 43.6 / 66.0& 29.2 / 44.9& \textbf{20.4} / 27.1& \textbf{28.4} / \textbf{42.5}& 41.1 / 57.5\\
\hline
    \end{tabular}
    \caption{Metrics of \textbf{\emph{Success Rate(\%) / SPL(\textperthousand)}} evaluating the performances of BRM and baselines agents using the ground truth oracle semantic signals provided by the environments. We also include the performance of the original BRM agent using CNN detector as a reference. The performance of BRM-CNN agents is comparable to BRM-true agents and sometimes even better. More discussions are in Sec.~\ref{sec:oracle}.} \label{fig:oracle}
\end{table*}

\section{Additional Results on Episode Length}\label{app:length}
We illustrate the ground truth shortest distance information as well as the average episode length of success episodes for all the approaches. The results are shown in Table~\ref{fig:epl}. The average ground truth shortest path is around 46.86 steps. Note that the agent has 9 actions per step and suffers from strong partial observability, which indicates the difficulty of the task.

\begin{table*}[h]
	\centering
	\begin{tabular}{c|c|c|c|c|c|c}
		\hline
		\hline
		\multicolumn{7}{c}{Average Ground Truth Shortest Path Length}\\
		\hline
		\hline
		plan-dist&1&2&3&4&5&overall\\
		\hline
		\hline
		Oracle & 12.27&42.53&61.09&72.47&63.74&46.86\\
		\hline
		\hline
		\multicolumn{7}{c}{Average Successful Episode Length}\\
		\hline
		\hline
		plan-dist&1&2&3&4&5&overall\\
		\hline
		\hline
		\multicolumn{7}{c}{Horizon $H=300$}\\
		\hline
		\hline
random& 34.0& 112.7& 143.8& 148.0& 149.7& 89.8\\
\hline
pure $\mu(\theta)$& 55.2& 107.0& 127.9& 140.8& 139.4& 84.7\\
\hline
aug.$\mu_S(\theta)$& 49.7& 112.5& 159.9& 179.1& 176.8& 89.2\\
\hline
RNN control.& 65.0& 132.3& 157.2& 142.7& 144.1& 111.8\\
\hline
BRM& 56.5& 124.4& 167.8& 150.7& 127.6& 107.7\\
\hline
		\hline
		\multicolumn{7}{c}{Horizon $H=1000$}\\
		\hline
		\hline
random& 121.7& 354.7& 426.6& 532.8& 409.5& 322.1\\
\hline
pure $\mu(\theta)$& 55.2& 107.0& 127.9& 140.8& 139.4& 84.7\\
\hline
aug.$\mu_S(\theta)$& 163.1& 360.9& 471.9& 460.7& 432.5& 307.1\\
\hline
RNN control.& 174.0& 368.4& 465.3& 466.6& 397.6& 339.5\\
\hline
BRM& 172.9& 350.5& 460.0& 512.3& 418.1& 337.0\\
\hline
	\end{tabular}
	\caption{Averaged successful episode length for different approaches. The length of shortest path reflects the strong difficulty of this task.} \label{fig:epl}
\end{table*}

\section{Additional Implementation Details}
The source code is available at \url{https://github.com/jxwuyi/HouseNavAgent}.

\subsection{Learning the LSTM Locomotion}\label{app:policy}
\textbf{Policy Architecture: }We utilize the same policy architecture and settings as \cite{wu2018building}: we have 4 convolution layers of 64, 64, 128, 128 channels each and with kernel size 5 and stride 2, an MLP layer of 256 units, an LSTM cell of 256 units, two MLP layers of 126 and 64 units for policy head and another 2 MLP layers of 64 and 32 units for value head. Batch normalization is applied to all the layers before LSTM. Activation is ReLU. The a only difference is that the original policy uses a gated attention mechanism for target conditioning while we use a behavior approach by training a separate sub-policy for each semantic target.

For the semantic augmented policy, we feed the semantic information to the MLP layer before LSTM.

\textbf{Hyperparameters: } 
We run a parallel version of A2C~\cite{mnih2016asynchronous} with 1 optimizer and 200 parallel rollout workers, each of which simulates a particular training house. We collect a training batch of 64 trajectories with 30 continuous time steps in each iteration. We set $\gamma=0.97$, batch size 64, learning rate 0.001 with Adam, weight decay $10^{-5}$ and entropy bonus 0.1. We also add the squared $l_2$ norm of policy logits to the total loss with a coefficient of 0.01. We normalize the advantage to mean 0 and standard deviation 1. We totally run 60000 training iterations and use the final model as our learned policy.

\textbf{Reward shaping: } The reward at each time step is computed by the difference of shortest paths in meters from the agent's location to the goal after taking a action. We also add a time penalty of 0.1 and a collision penalty of 0.3. When the agent reaches the goal, the success reward is 10.

\textbf{Curriculum learning: }We run a curriculum learning by increasing the maximum of distance between agent's birth meters and target by 3 meters every 10000 iterations. We totally run 60000 training iterations and use the final model as our learned policy $\mu(\theta)$.

\subsection{Building the Relational Graph}\label{app:model}
We run random exploration for 300 steps to collect a sample of $z$. For a particular environment, we collect totally 50 samples for each $z_{i,j}$. 
For all $i\ne j$, we set $\psi^{\mathrm{obs}}_{i,j,0}=0.001$ and $\psi^{\mathrm{obs}}_{i,j,1}=0.15$.

\subsection{Training the CNN Semantic Extractor}\label{app:cnn}
We take the panoramic view as input, which consists of 4 images, $s_o^{1},\ldots,s_o^{4}$ with different first person view angles. The only exception is that for target ``outdoor'', we notice that instead of using a panoramic view, simply keeping the recent 4 frames in the trajectory leads to the best prediction accuracy. We use an CNN feature extractor to extract features $f(s_o^i)$ by applying CNN layers with kernel size 3, strides $[1, 1, 1,  2,  1, 2, 1, 2, 1, 2]$ and channels $[4, 8, 16, 16, 32, 32, 64, 64, 128, 256]$. We also use relu activation and batch norm. 
Then we compute the attention weights over these 4 visual features by $l_i=f(s_o^i)W_1^TW_2\left[f(s_o^1),\ldots,f(s_o^4)\right]$ and $a_i=\textrm{softmax}(l_i)$. Then we compute the weighted average of these four frames $g=\sum_i a_i f(s_o^i)$ and feed it to a single layer perceptron with 32 hidden units. For each semantic signal, we generate 15k positive and 15k negative training data from $\mathcal{E}_{\textrm{train}}$ and use Adam optimizer with learning rate $5e$-$4$, weight decay $1e$-$5$, batch size 256 and gradient clip of 5. We keep the model that has the best prediction accuracy on $\mathcal{E}_{\textrm{valid}}$.

For a smooth prediction during testing, we also have a hard threshold and filtering process on the CNN outputs: $s_s(T_i)$ will be 1 only if the output of CNN remains a confidence for $T_i$ over 0.9 for consecutively 3 steps.

\end{document}